\documentclass[remotesensing,submit,pdftex,moreauthors]{Definitions/mdpi} 
\let\linenumbers\relax
\firstpage{1} 
\pubvolume{1}
\issuenum{1}
\articlenumber{0}
\pubyear{2024}
\copyrightyear{2024}
\hreflink{https://doi.org/10.48550/arXiv.2401\linebreak
01375}
\doinum{10.48550/arXiv.2401.01375}
\usepackage[T1]{fontenc}
\usepackage{subfig}
\usepackage{siunitx}
\usepackage{array}
\usepackage{fancyhdr}
\usepackage{adjustbox}

\Title{Mapping Walnut Water Stress with High Resolution Multispectral UAV Imagery and Machine Learning}
\TitleCitation{Mapping Walnut Water Stress with High Resolution Multispectral UAV Imagery and Machine Learning}
\Author{Kaitlyn Wang $^{1,\dagger,\ddagger}$\orcidA{} and Yufang Jin $^{2,}$*}

\AuthorNames{Kaitlyn Wang and Yufang Jin}

\AuthorCitation{Wang, K.; Jin, Y.}
\address{%
$^{1}$ \quad The Harker School, 500 Saratoga Ave, San Jose, CA 95129, USA; 24kaitlynw@students.harker.org\\
$^{2}$ \quad Dept of Land, Air and Water Resources, UC Davis, CA 95616, USA; yujin@ucdavis.edu}

\corres{Correspondence: 24kaitlynw@students.harker.org}

\abstract{Effective monitoring of walnut water status and stress level across the whole orchard is an essential step towards precision irrigation management of walnuts, a significant crop in California. This study presents a machine learning approach using Random Forest (RF) models to map stem water potential (SWP) by integrating high-resolution multispectral remote sensing imagery from Unmanned Aerial Vehicle (UAV) flights with weather data. From 2017 to 2018, five flights of an UAV equipped with a seven-band multispectral camera were conducted over a commercial walnut orchard, paired with concurrent ground measurements of sampled walnut plants. The RF regression model, utilizing vegetation indices derived from orthomosaiced UAV imagery and weather data, effectively estimated ground-measured SWPs, achieving an $R^{2}$ of 0.70 and a mean absolute error (MAE) of 0.80 bars. The integration of weather data was particularly crucial for consolidating data across various flight dates. Significant variables for SWP estimation included wind speed and vegetation indices such as NDVI, NDRE, and PSRI. As a comparison, a reduced RF model excluding red-edge indices of NDRE and PSRI, demonstrated slightly reduced accuracy ($R^{2}$ = 0.63). Subsequently, a random-forest classification model was developed to categorize the severity of walnut plant water stress to three distinct stress levels: low stress, moderate stress, and severe stress.The RF classification model predicted water stress levels in walnut trees with 85\% accuracy, surpassing the 80\% accuracy of the reduced classification model. The results affirm the efficacy of UAV-based multispectral imaging combined with machine learning, incorporating thermal data, NDVI, red-edge indices, and weather data (wind, temperature, VPD), in walnut water stress estimation and assessment. This methodology enables a scalable, cost-effective tool for data-driven precision irrigation management at an individual plant level in walnut orchards. It highlights the promising capability to replace manual, labor-intensive measurements with precise, targeted irrigation, marking a significant advancement in refining and optimizing irrigation strategies.
}


\begin{document}
%\end{document}
\section{Introduction}

California's Central Valley is a pivotal region for walnut production, contributing over 99\% of the walnuts grown in the U.S. and supplying about half of the global walnut trade. This makes walnuts one of the most economically significant specialty crops in California, with an annual retail value exceeding \$15.4 billion (\href{https://www.federalregister.gov/documents/2023/10/27/2023-23729/walnuts-grown-in-california-decreased-assessment-rate}{Federal Register 2023}). However, the region's recent severe drought conditions have compelled growers to rely heavily on groundwater pumping, leading to alarmingly rapid rates of groundwater depletion \citep{Famiglietti_2014}. In response to these challenges, the optimization of water management practices has become a critical necessity for crop producers, especially given the escalating water demands tied to the expansion of specialty crops in the Central Valley.
Timely and accurate mapping of water status across entire blocks is essential for implementing adaptive and precision irrigation management at both zonal and individual walnut plant levels, thereby enhancing water use efficiency. This is particularly crucial when initiating irrigation \citep{Intrigliolo_Castel_2007}. Conventionally, walnut plant physiological water status is gauged through stem and/or leaf water potentials using pressure chambers \citep{Chone_2001, Williams_2002}. However, this approach, being labor-intensive and typically limited to a small number of sample walnut plants, does not effectively support the mapping of within-field variability or the monitoring of temporal changes in orchard blocks.

Conversely, remote sensing techniques, such as unmanned aerial vehicles (UAVs) and satellite-based imaging, offer promising and cost-effective alternatives for mapping walnut water stress levels. Notably, several small-scale UAV studies employing multispectral imagery have demonstrated significant correlations between various infrared and red-edge vegetation indices (VIs) and plant water stress levels \citep{Baluja_2012, Espinoza_2017, Zhang_2019}. These methods hold the potential for large-scale, efficient monitoring of walnut health and irrigation needs.

Meanwhile, various machine learning techniques have been applied to predict water stress dynamics in crops \citep{Romero_2018}, with some advanced studies incorporating weather data to improve accuracy \citep{Tang_Jin_2023, Tang_Jin_2022}. Deep learning methods, including convolutional neural networks (CNN) and Inception-Resnet, have been used to classify multispectral imaging (MSI) of agricultural fields. These methods leverage pixelwise spectral data, spectral-spatial features, and sometimes additional principal components \citep{Hsieh_2020, Zhang_2019}. Typically, deep learning models require thousands of training samples, and various data augmentation techniques for these models are discussed in \cite{Illarionova_2021}.

In our study, given the limited number of data samples available (a total of 200), a deep learning approach was deemed unsuitable due to its typically high data requirements. Instead, we opted for machine learning methods like Random Forest, which are well-suited for achieving reliable results with smaller datasets. Consequently, this approach was selected for our analysis.

We focus on analyzing variables derived from UAV multispectral imagery alongside weather factors that significantly influence walnut stem water potential (SWP). Our goal is to develop a robust model capable of predicting walnut water stress, thereby substantially aiding in the implementation of precision irrigation practices.

The structure of this paper is as follows: Section 2 provides an in-depth view of our experimental design, detailing the models employed for both estimating and classifying plant water stress. Section 3 elaborates on the outcomes of our stem water potential (SWP) estimation, emphasizing the significance of various input variables. In Section 4, we present the findings of our walnut water stress classification. Section 5 delves into a comparative analysis of the models used, discusses the uncertainties in our results and examines the relationship between SWP and a range of input variables. The paper concludes with Section 6, which summarizes our research and proposes directions for future studies and potential areas for further investigation.

\section{Materials and Methods}
\subsection{Site Description and Experiment Design}

This study was conducted in a walnut orchard located near Davis, California, characterized by a typical Mediterranean climate with warm and dry conditions during the spring and summer growing seasons. The selected site comprises primarily coarse loamy sand and silt soils, spread across a uniformly flat terrain. The field benefits from a subsurface drip irrigation system.

\begin{figure}[ht]
  \centering
  \begin{minipage}[t]{0.42\textwidth}
    \centering
    \subfloat{\includegraphics[width=\textwidth]{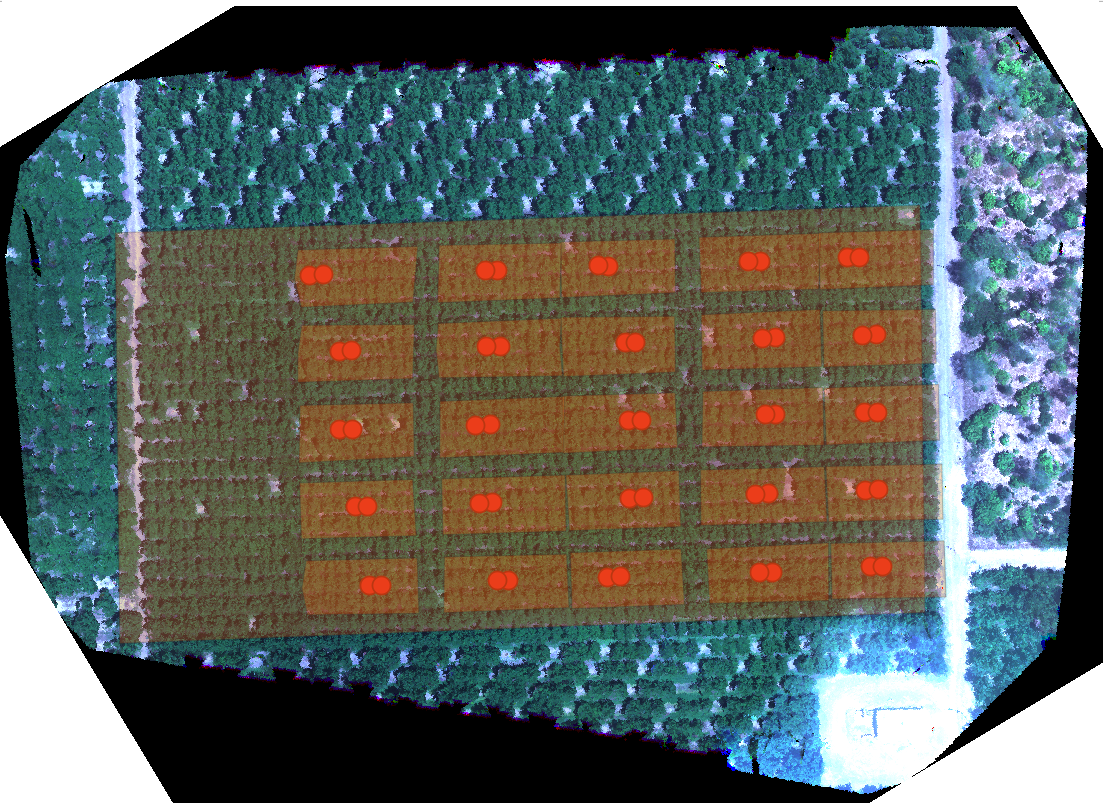}}\label{fig:field_image}
  \end{minipage}
  %\hfill
  \hspace{5mm}
  \begin{minipage}[t]{0.50\textwidth}
    \centering
    \subfloat{\includegraphics[width=\textwidth]{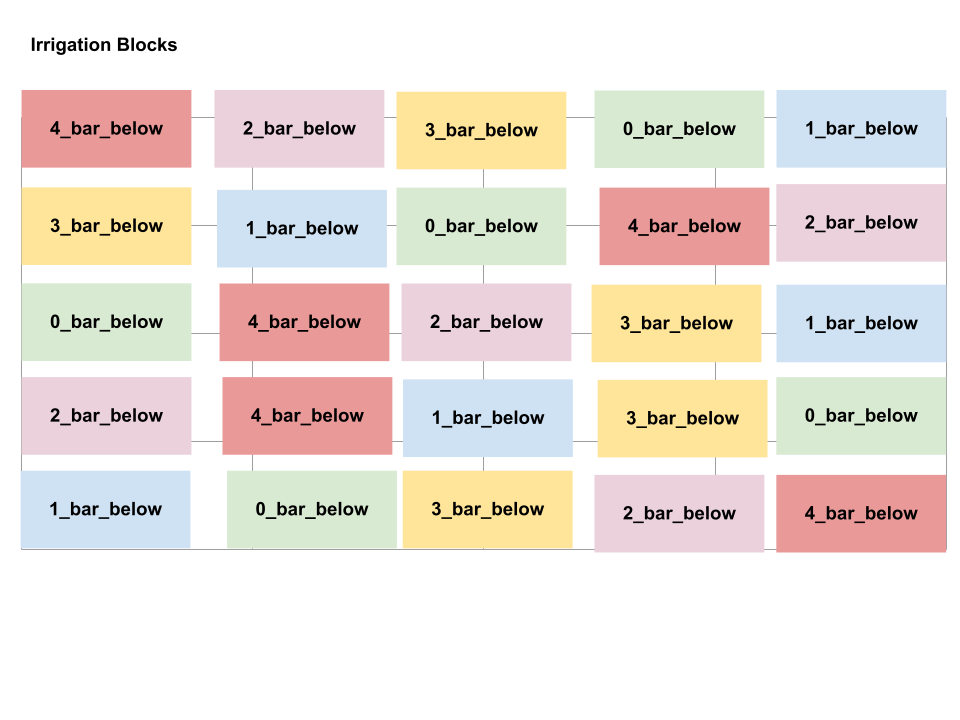}}\label{fig:block_irrig_treatment}
  \end{minipage}
  \caption{The study area (inset) and experimental layout of a mature commercial walnut orchard near UC Davis, California, are illustrated in the false color UAV imagery taken on July 09, 2018 (the upper plot). The orchard is segmented into twenty-five blocks, with their boundaries marked as orange rectangles on the UAV imagery. The locations of sampled walnut plants, where ground measurements were conducted, are indicated by red dots. The lower plot provides a detailed illustration of the irrigation treatment assigned to each block.}
  \label{fig:walnut_field}
\end{figure}

The walnut orchard under investigation is depicted in Figure \ref{fig:walnut_field}. For the purposes of this experiment, the orchard was segmented into twenty-five substantial blocks. Each block is systematically arranged in a block design. In the field image, these blocks are visually distinguished as orange rectangular shapes. Every block contains 27 walnut trees.

Initially, all plots in this orchard received full watering, allowing the walnut trees to achieve complete canopy cover. Since 2017, controlled irrigation treatments have been applied at the block level, as indicated in the lower plot of Figure \ref{fig:walnut_field}. Specifically, five different irrigation treatments were administered to each block, set at $X$ bar below the default level, where $X \in \{0, 1, 2, 3, 4\}$.

\begin{figure}
    \centering
    \includegraphics[scale = 0.47]{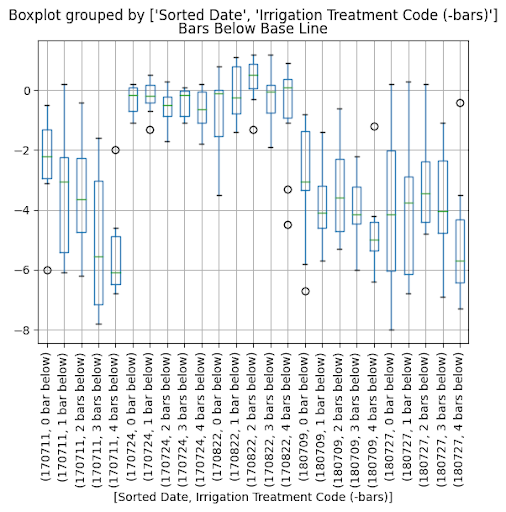}
    \caption{Daily measured SWP statistics across the five irrigation treatments. Each treatment is characterized by a reduction in irrigation intensity, set at $X$ bar below the standard level, with $X$ belonging to the set $\{0, 1, 2, 3, 4\}$. This graph provides a comparative analysis of the soil water potential under varying irrigation conditions.}
    \label{fig:swp_bar_plot}
\end{figure}

\begin{figure}
    \centering
    \includegraphics[scale = 0.45]{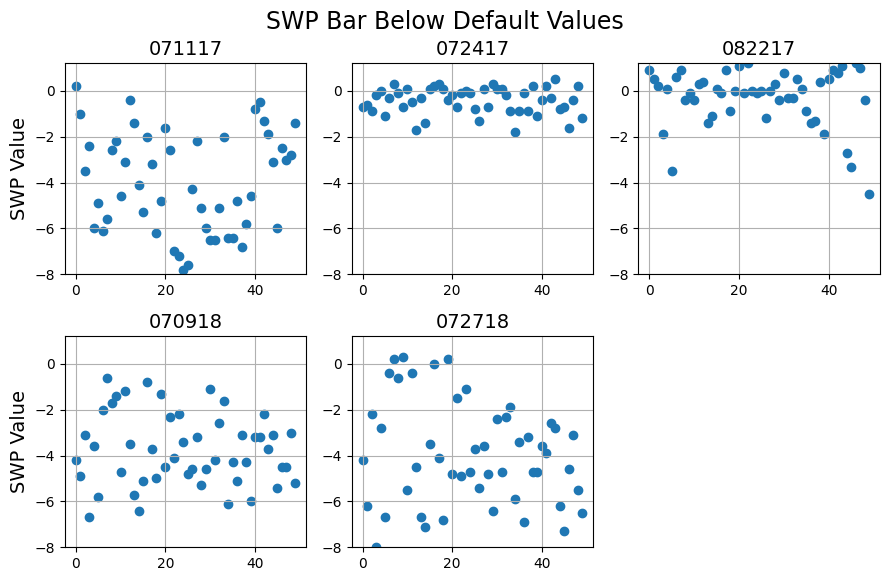}
    \caption{Variations in SWP values across different flight dates. This figure illustrates the diverse patterns in SWP value distributions observed on the five flight dates. These variations can be attributed to fluctuating environmental conditions, including changes in wind, temperature, and humidity, which significantly impact the SWP readings.}
    \label{fig:swp_all_dates}
\end{figure}

Over the course of early June to August in both 2017 and 2018, five Intensive Observation Periods (IOPs) were carried out. During these periods, extensive ground measurements were conducted using pressure chambers, complemented by aerial imagery captured through a UAV system. A series of five flights were executed over the study orchard, as documented in Table \ref{tab:ground_measure_date_weather}. Figure \ref{fig:swp_bar_plot} presents a detailed visualization of the SWP measurements corresponding to each irrigation treatment, captured on each of the five days when UAV flights were conducted. 
Figure \ref{fig:swp_all_dates} presents the measured SWP values across various flight dates, showcasing distinct patterns in SWP value distributions on each of the five dates. These fluctuations in the data can be linked to varying environmental conditions, such as alterations in wind speed, ambient temperature, humidity, and VPD levels, all of which markedly influence the SWP measurements.

\subsection{UAV Flights and Image Processing}
We acquired multispectral aerial imagery using a MicaSense RedEdge camera attached to a DJI Matrice 100 quadcopter, during the months of July and August in 2017 and 2018. The MicaSense RedEdge camera (MicaSense Inc., Washington, USA) is equipped with seven spectral bands that capture reflected radiation in the blue (centered at 475 nm), green (560 nm), panchromatic (634 nm), red (668 nm), red-edge (717 nm), near-infrared (840 nm), and long-wave infrared (LWIR) (\SI{11}{\um}) spectra.

A series of five flights were executed over the study orchard, as documented in Table \ref{tab:ground_measure_date_weather}. The UAV was operated at a height of 120 meters above ground level, yielding an image resolution of 8 cm. To mitigate shadow effects, all flights were scheduled around solar noon. Radiometric calibration was ensured by capturing images of a calibrated reflectance panel immediately before and after each flight. For accurate georeferencing, five ground control points (GCPs) were strategically placed in each orchard: four at the corners and one at the center. The GPS coordinates of these GCPs were accurately recorded using a Trimble Geo 7x device (Trimble, CA, USA). 

\begin{table}
\centering
\caption{Summary of UAV imagery and corresponding ground measurements of stem water potential during Intensive Observation Periods (IOPs) in the walnut orchard. The table also includes data on the noon-time weather conditions concurrent with these measurements.}
\begin{tabular}{ p{2.0cm} p{3.0cm} p{2.5cm} p{1.3cm} p{1.5cm} p{1.5cm}
}
 \hline
 Date & Number of ground measurements & Air temperature ($^{\circ}$F) & Humidity (\%) & Wind (mph) & VPD (kPa)\\
 \hline
07/11/2017 & 39 & 90.55 & 31.5 & 5.0 & 3.355\\
07/24/2017 & 50 & 90.80 & 32.5 & 16.0 & 3.332\\
08/22/2017 & 50 & 83.30 & 58.5 & 5.0 & 1.615\\
07/09/2018 & 50 & 90.20 & 28.7 & 2.2 & 3.455\\
07/27/2018 & 50 & 89.54 & 49.3 & 2.9 & 2.404\\
 \hline
\end{tabular}
\label{tab:ground_measure_date_weather}
\end{table}

The raw multispectral images were stitched and processed using Pix4DMapper Pro photogrammetry software (Pix4D LLC, Switzerland). Initially, the software automatically generated tie points. Subsequently, the coordinates of the GCPs were imported, enhancing geolocation accuracy to within a 30 cm error margin. In the next phase, a point cloud and 3D textured mesh were constructed from these tie points, leading to the development of a digital surface model (DSM). Radiometric calibration was carried out using data from the calibration panel, enabling the production of orthomosaics for reflectance maps in each spectral band with a resolution of 8 cm. To ensure spatial consistency among images from multiple dates, image co-registration was conducted. Specifically, co-registration techniques in QGIS were used to align all walnut UAV images to the spatial coordinates from the imagery of the final flight date.

\subsection{Field Measurements}
In each of the twenty-five blocks within the orchard, biophysical measurements were systematically conducted on two adjacent walnut plants. These stem water potential (SWP) measurements were taken in tandem with the UAV flights, employing pressure chambers (PMS Inc., Oregon, USA) within a one-hour window centered around local solar noon. The specifics of these measurements, including the number of sampled walnuts coinciding with each UAV flight for every Intensive Observation Period (IOP), are detailed in Table \ref{tab:ground_measure_date_weather}, which also includes relevant weather conditions.

Hourly weather data, encompassing air temperature and relative humidity, were obtained from the UCD/NOAA Climate Station website (\url{https://atm.ucdavis.edu/weather/uc-davis-weather-climate-station}). The station, situated west of the UC Davis campus, was selected as the primary source for meteorological data pertinent to the orchard's location.

\subsection{Data Preprocessing: Canopy Segmentation}
In order to differentiate walnut canopy from soil and canopy shadows in the UAV imagery, we employed a dual-mask approach combining the Digital Surface Model (DSM) and the Normalized Excess Green (NExG) index for each flight. The DSM, generated by the photogrammetry software Pix4DMapper, captures the earth's surface topography, encompassing all objects such as trees and buildings. It serves as a static layer representing the structural information of the trees. 

The NExG index, on the other hand, effectively contrasts the green part of the spectrum against red and blue, making it a valuable tool for distinguishing vegetation from non-vegetative elements and discerning canopy shadows within the visible spectrum. The NExG index is calculated as per the following equation:
\begin{equation}\label{NExG_equation}
 NExG = \frac{2\times{Green}-Red-Black}{Green+Red+Black},
\end{equation}
where 'Green', 'Red', and 'Black' refer to the respective spectral band values in the multispectral imagery. This method enables the precise segmentation of canopy areas, facilitating more accurate and targeted analysis of the walnut plant water pressure.

\begin{figure}
  \centering
  \begin{minipage}[c]{0.42\textwidth}
    \centering
    \subfloat{\includegraphics[width=\textwidth]{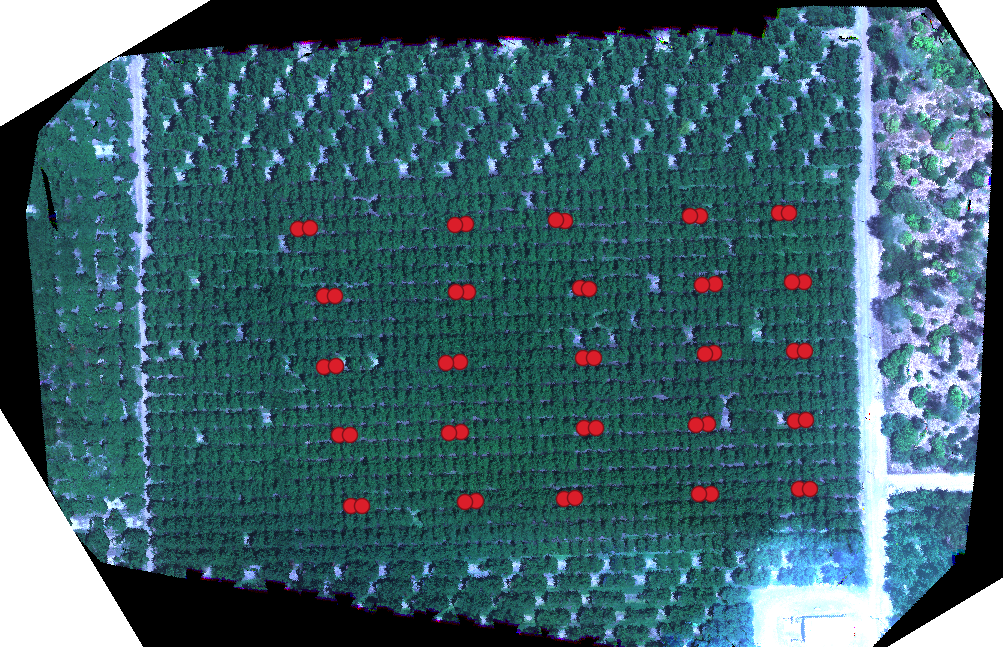}}\label{fig:field_image_no_block}
  \end{minipage}
  %\hfill
  \hspace{2mm}
 % Add horizontal space here
  \vspace{0.2cm} % Adjust the 1cm to the desired amount of space  
  \begin{minipage}[c]{0.45\textwidth}
    \centering
    \subfloat{\includegraphics[width=\textwidth]{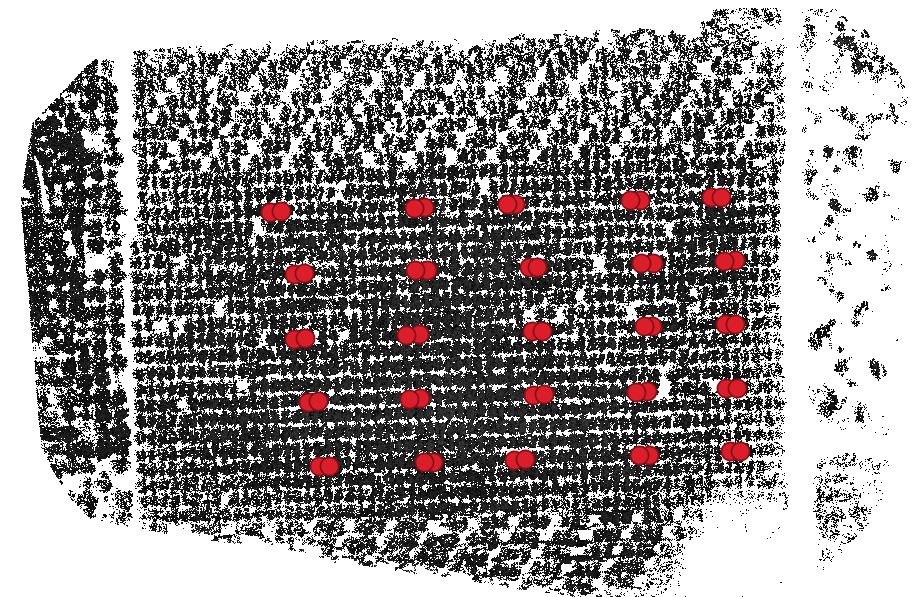}}\label{fig:canopy_dsm_ExG}
  \end{minipage}
  \caption{Comparison of UAV imagery before and after the application of the DSM and NExG non-canopy masks, illustrated in the upper and lower images, respectively.}
  \label{fig:canopy_dsm_nexg}
\end{figure}

\begin{figure}
  \centering
  \begin{minipage}[c]{0.28\textwidth}
    \centering
    \subfloat{\includegraphics[width=\textwidth]{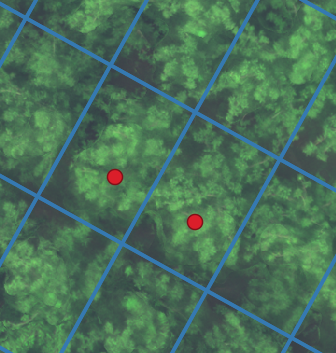}}\label{fig:grid_canopy_1}
  \end{minipage}
  %\hfill
  \hspace{14mm}
  \begin{minipage}[c]{0.28\textwidth}
    \centering
    \subfloat{\includegraphics[width=\textwidth]{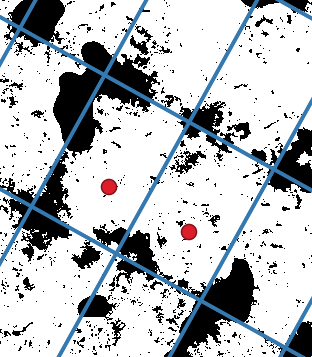}}\label{fig:grid_canopy_2}
  \end{minipage}
  \caption{A walnut field segmented into grids, each approximately encompassing a single canopy. The upper image shows UAV imagery of the field with grids overlaid, while the lower image displays the same area after the application of the DSM and NExG non-canopy masks, illustrating the effect of these masks on canopy identification.}
  \label{fig:grid_canopy}
\end{figure}

Both the DSM and NExG histograms over all pixels showed a bimodal distribution, one representing plant canopy and the other representing soil background. We identify a vegetated pixel if both DSM and NExG identify it as vegetation. The soil background pixels were then masked out, and only the remaining vegetated pixels were used for further analysis (Figure \ref{fig:canopy_dsm_nexg}). Visual inspection was conducted in the QGIS software to confirm the effectiveness of the canopy segmentation from background soil and shadows. We further applied a $100 \times{100} (4.5m \times{4.5m})$ grid size over the whole orchard, with each grid containing approximately one walnut canopy; the median vegetation indices within the grid were then extracted to match the canopy of each walnut plant in the orchard (Figure \ref{fig:grid_canopy}.

The subsequent step in our data processing involved overlaying a grid with dimensions of $100 \times 100$ (equivalent to $4.5m \times 4.5m$) across the entire orchard. This grid resolution was chosen to approximately cover the area of a single walnut canopy within each grid cell. For each cell, we then extracted the median values of the vegetation indices. These median values were aligned with the respective canopy of each walnut plant in the orchard. This process and its results are visually demonstrated in Figure \ref{fig:grid_canopy}, providing a clear representation of the spatial distribution of vegetation features relative to individual canopies.

%\subsection{Our Approaches for Stem Water Potential Estimation}
\subsection{Selection of Vegetation Indices}

Vegetation indices (VIs) derived from UAV multispectral imagery are pivotal in the realm of precision agriculture. These indices capture a range of plant physiological attributes, including canopy structure, chlorophyll content, xanthophyll cycle activity, and sun-induced fluorescence, thereby serving a multitude of agricultural applications. Their effectiveness spans from crop mapping \citep{Huang_2018}, to monitoring plant nitrogen levels \citep{Barnes_2000}, and identifying water stress in plants \citep{Suarez_2008, Berni_2009, Baluja_2012, Zarco_2013}. The careful selection of relevant VIs is critical for enhancing the accuracy of estimating plant stem water potential, a key factor in efficient agricultural management and water stress detection.

In our study, we computed several widely-used VIs using reflectance data from different bands in the orthomosaic UAV imagery, as detailed in Table \ref{tab:veg_indices}. Plant stress often causes an increase in visible reflectance, a decrease in chlorophyll and absorption of visible light. Additionally, reduced near-infrared (NIR) reflectance will happen due to changes in the leaf or stem tissue. The most used spectrum bands are Green, Red, Red-edge, and NIR \citep{Gao_review_2020}. The NIR band Normalized Difference Vegetation Index (NDVI) is notably the most prevalent index for assessing plant growth, vigor, and structure \citep{Rouse_1974}. Studies have established a significant correlation between NDVI and overall plant stress levels, particularly due to the cumulative effects of water deficit on plant growth \citep{Acevedo_2008, Baluja_2012, Poblete_2017}. However, NDVI and similar spectral indices exhibit a delayed response to the initial onset of vegetation stress. In contrast, approaches based on thermal-infrared data can promptly detect plant water stress, provided that orchard remote sensing imagery is available \citep{Knipper_2019}. \cite{Zhang_2019} found TCARI, RDVI and SAVI had the best correlation with CWSI (crop water stress index). 
In our study, we computed several widely-used VIs using reflectance data from different bands in the orthomosaic UAV imagery, as detailed in Table \ref{tab:veg_indices}. Plant stress often causes an increase in visible reflectance, a decrease in chlorophyll and absorption of visible light. Additionally, reduced near-infrared (NIR) reflectance will happen due to changes in the leaf or stem tissue. The most used band channels are Green, Red, Red-edge and NIR \citep{}. The Normalized Difference Vegetation Index (NDVI) is notably the most prevalent index for assessing plant growth, vigor, and structure \citep{Rouse_1974}. Studies have established a significant correlation between NDVI and overall plant stress levels, particularly due to the cumulative effects of water deficit on plant growth \citep{Acevedo_2008, Baluja_2012, Poblete_2017}. However, NDVI and similar spectral indices exhibit a delayed response to the initial onset of vegetation stress. In contrast, approaches based on thermal-infrared data can promptly detect plant water stress, provided that orchard remote sensing imagery is available \citep{Knipper_2019}. \cite{Zhang_2019} found TCARI, RDVI and SAVI had the best correlation with CWSI (crop water stress index).

Additionally, research by \cite{Tang_Jin_2022}, along with earlier studies \citep{Tilling_2007}, suggests that reflectance from the red edge band indices like Normalized Difference Red Edge (NDRE), and the Plant Senescence Reflectance Index (PSRI) are effective indicators of canopy chlorophyll content.

Physiological research has established that weather conditions, such as air temperature, relative humidity, and vapor pressure deficit (VPD), play a crucial role in regulating plant stomatal closure and atmospheric water exchange, thereby directly influencing plant water status \citep{Garnier_1985}. In light of this, models that integrate multiple VIs from UAV imagery with weather variables have shown promising results in monitoring plant water stress. An example by \cite{Tang_Jin_2022} successfully incorporated weather data into such a model, achieving accuracy in plant water stress prediction.

For a detailed summary of the vegetation indices employed in this study, please see Table \ref{tab:veg_indices} below. The table enumerates each index, providing its formula and the corresponding reference for further context.

% Adjust the row height
\renewcommand{\arraystretch}{1.6} % Adjust the 1.5 as needed
\begin{table*}
%\centering
\caption{Vegetation indices used for this study}
\begin{tabular}{ p{5.4cm} p{1.9cm} p{3.7cm} p{1.5cm}
}
 \hline
 Vegetation index & Abbreviation & Formula & Reference\\
 \hline
Structural indices & & &\\
Normalized difference vegetation index & NDVI & $NDVI = \frac{NIR-Red}{NIR+Red}$ & \citep{Rouse_1974}\\
Chlorophyll indices & & &\\
Normalized difference red edge & NDRE & $NDRE = \frac{NIR-Rededge}{NIR+Rededge}$ & \citep{Barnes_2000}\\
Plant senescence reflectance index & PSRI & $PSRI = \frac{Red-Blue}{Rededge}$ & \citep{Merzlyak_1999}\\

 \hline
\end{tabular}
\label{tab:veg_indices}
\end{table*}
% Reset the row height for subsequent tables
\renewcommand{\arraystretch}{1} % Resets to default

\subsection{Statistical Approaches}

\begin{figure*}
%  \begin{minipage}[c]{0.9\textwidth} % Adjust width as needed
    \centering
    \subfloat{\includegraphics[width=\textwidth]{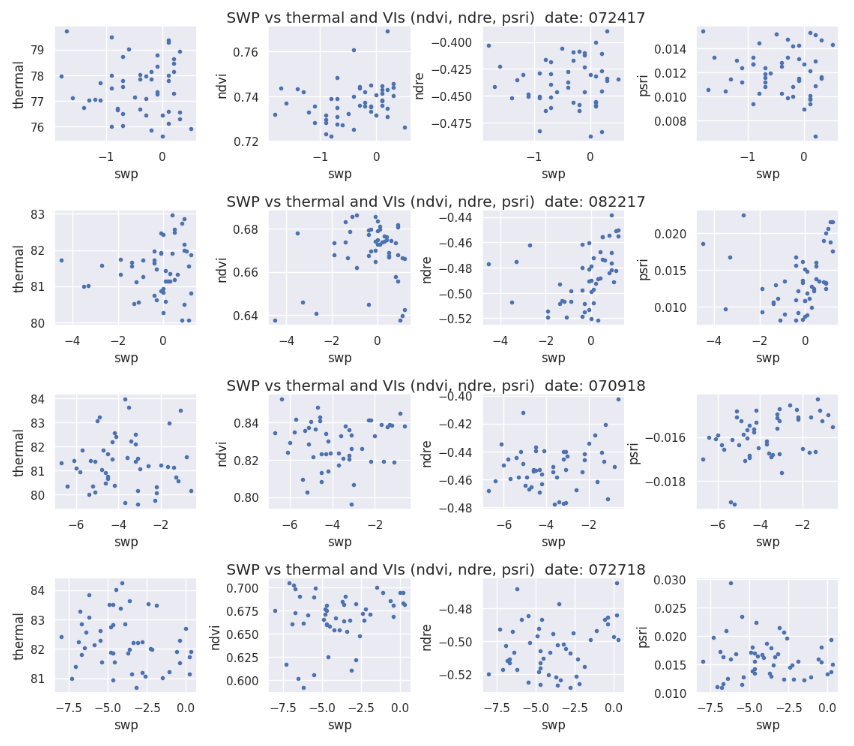}}\label{fig:correlation_swp_thermal_VIs}
  \caption{Analysis of Univariable Correlations: Field-Measured Soil Water Potentials (SWPs) Versus Thermal Values and Various Vegetation Indices}
  \label{fig:correlation_univar}
\end{figure*}

Our initial analysis involved exploring the univariate relationships between field-measured stem water potentials (SWPs) and thermal values, as well as various vegetation indices. The results of this examination are presented in Figure \ref{fig:correlation_univar}. It was observed that there is no discernible linear correlation between SWP and any individual variable. Furthermore, the SWP data indicated significant variations in weather conditions across different flight dates, implying that SWPs measured on distinct dates are not directly comparable.

Given these initial findings, we employed machine learning techniques to delve deeper into the multivariable relationships between field-measured SWPs and a spectrum of input variables. This approach was designed to encompass both temporal and spatial variances, providing a view of the interplay between SWP and an array of environmental and spectral factors.

Machine learning algorithms are highly effective at identifying patterns and correlations within large and complex datasets, making them particularly suitable for this kind of analysis. In our study, we developed both regression and classification models to evaluate the predictive capability of various variables on stem water potential (SWP). These models were trained and tested on the dataset comprising UAV-derived vegetation indices, thermal-band values from the UAV imagery, and relevant weather parameters, in addition to the ground-measured SWP values. Our goal was to create a predictive model capable of accurately estimating both SWP and water stress levels in walnut orchards, utilizing readily accessible data from UAVs and weather observations.

\subsection{Machine Learning Approaches}

The Random Forest (RF) algorithm, a prominent machine learning technique, has gained widespread use in remote sensing applications, such as biomass estimation \citep{Zhou_2016} and surface temperature retrieval \citep{Yang_2017}. In this study, we developed multiple RF models to explore the potential of utilizing UAV multispectral imagery in conjunction with weather data to estimate walnut plant water stress. These models were developed, verified, and tested using ground-measured Soil Water Potentials (SWPs) of walnut trees within the orchard. Our comprehensive RF model integrated three distinct categories of predictors: thermal-band values derived from UAV imagery, a suite of vegetation indices (NDVI, NDRE, PSRI), and a set of weather parameters (air temperature, VPD, and wind speed).

In this study, data from the initial UAV flight on 07/11/2017 were excluded due to suboptimal quality. The data gathered in the subsequent four field observations, totaling 200 samples, were fed to an RF model and divided using a random split of 80\%-10\%-10\%. Specifically, 80\% of the samples (n=160) constituted the training set, while 10\% (n=20) were allocated for validation, and the remaining 10\% (n=20) were designated as the independent testing set. Note that the division of data did not need to take into account the variation in flight dates or irrigation treatments. To further enhance the robustness of our model and ensure its reliability, a 10-fold cross-validation technique was employed. 

For assessing the performance of RF models, we employed three statistical metrics: the coefficient of determination ($R^{2}$), the root mean squared error (RMSE), and the mean absolute error (MAE). This evaluation process was iteratively conducted 10 times and the mean and standard deviations of the statistical measures were summarized from the results.

First, a Random Forest (RF) regression model was constructed to estimate plant stem water potential (SWP) values, utilizing the input variables mentioned earlier. This regression model is capable of predicting SWP values across the entire orchard, thereby enabling a comprehensive assessment of plant water stress throughout the orchard.

Subsequently, a Random Forest (RF) classification model was constructed to categorize the severity of plant water stress, an essential step toward implementing adaptive precision irrigation management. Utilizing the Walnut Stem Water Potential (SWP) Interpretation Guidelines (referenced at \href{https://www.sacvalleyorchards.com/manuals/stem-water-potential/pressure-chamber-advanced-interpretation-in-walnut/}{Walnut SWP Interpretation Guidelines}), we classified SWP values (measured in bars below baseline) into three distinct stress levels: low stress (\( swp \geq -0.4 \) bars), moderate stress (\( swp \in (-3, -0.4) \) bars), and severe stress (\( swp \leq -3 \) bars). These thresholds were determined based on our sample distribution, with an aim to maintain a relatively balanced number of samples across each category.

In order to assess the impact of various vegetation indices and weather variables on the accuracy of our plant water stress estimation models, additional reduced models were constructed using subsets of the initial input variables. For instance, recognizing that many low-cost sensors lack a red edge band, we tested a specific set of these reduced models by excluding both the red edge reflectance and all associated red-edge vegetation indices (VIs) from the predictor list. This 'NoRedEdge' model allowed us to evaluate the performance implications of omitting red-edge data in the context of low-cost sensor applications.

\section{SWP Estimation Results}
\subsection{Full RF Regression Model Based on Single Flight Date Data}

\begin{figure}
  \centering

  % All subplots in a single row with adjusted widths
  \begin{minipage}[c]{0.24\textwidth}
    \centering
    \subfloat{\includegraphics[width=\textwidth]{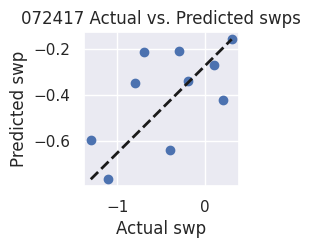}}\label{fig:rf_regr_072417}
  \end{minipage}
  \begin{minipage}[c]{0.24\textwidth}
    \centering
    \subfloat{\includegraphics[width=\textwidth]{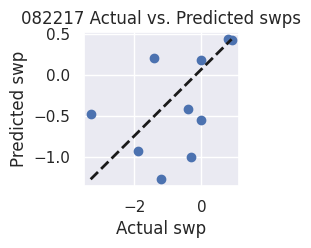}}\label{fig:rf_regr_082217}
  \end{minipage}
  \begin{minipage}[c]{0.24\textwidth}
    \centering
    \subfloat{\includegraphics[width=\textwidth]{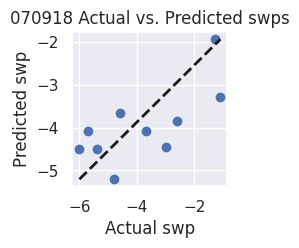}}\label{fig:rf_regr_070918}
  \end{minipage}
  \begin{minipage}[c]{0.24\textwidth}
    \centering
    \subfloat{\includegraphics[width=\textwidth]{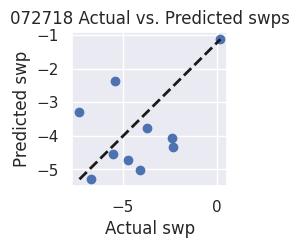}}\label{fig:rf_regr_072718}
  \end{minipage}

  \caption{Performance of the Full RF Regression Model: Predictions Based on Data from Individual Flight Dates}
  \label{fig:rf_regr_single_dates}
\end{figure}

\begin{figure}[ht]
    \centering
    \begin{minipage}[t]{0.48\linewidth}
        \centering
       \includegraphics[scale = 0.46]{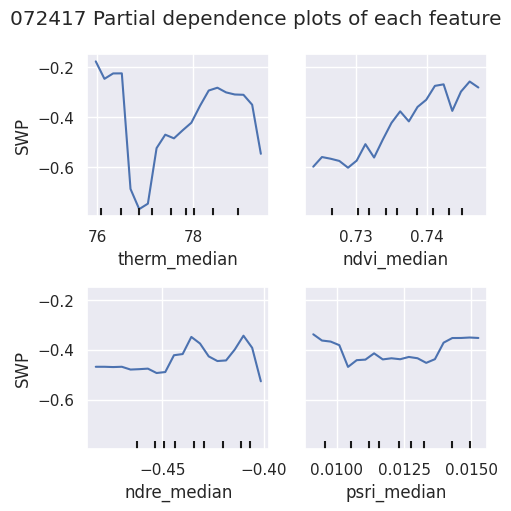}
        \caption{Partial dependence plots illustrating the influence of each feature on SWP measured on flight-date 072417.} 
        \label{fig:rf_regr_full_pdp_072417}
    \end{minipage}\hfill
    \begin{minipage}[t]{0.48\linewidth}
        \centering
        \includegraphics[scale = 0.46]{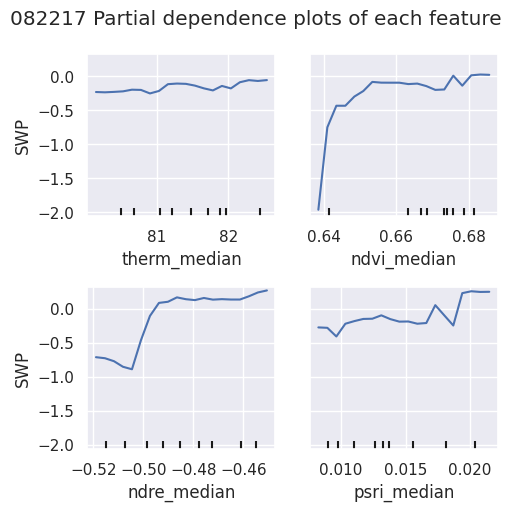}
        \caption{Partial dependence plots illustrating the influence of each feature on SWP measured on flight-date 082217.} 
        \label{fig:rf_regr_full_pdp_082217}
    \end{minipage}
\end{figure}

\begin{figure}[ht]
    \centering
    \begin{minipage}[t]{0.48\linewidth}
        \centering
        \includegraphics[scale = 0.46]{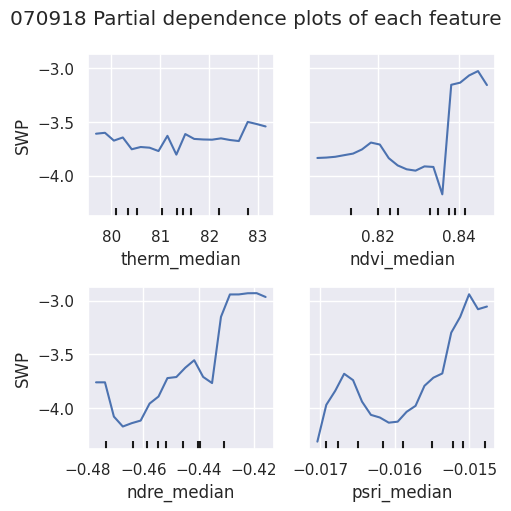}
        \caption{Partial dependence plots illustrating the influence of each feature on SWP measured on flight-date 070918.} 
        \label{fig:rf_regr_full_pdp_070918}
    \end{minipage}\hfill
    \begin{minipage}[t]{0.48\linewidth}
       \centering
        \includegraphics[scale = 0.46]{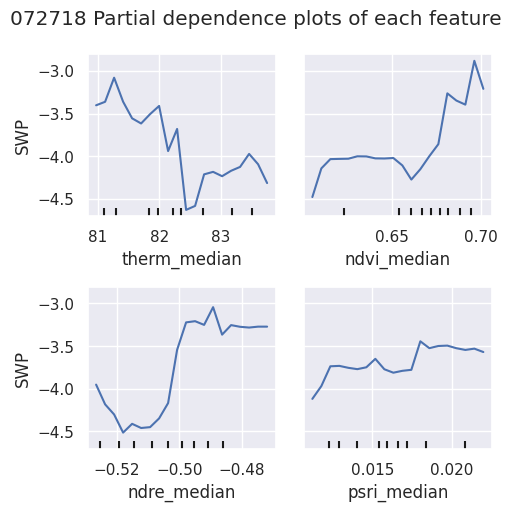}
        \caption{Partial dependence plots illustrating the influence of each feature on SWP measured on flight-date 072718.} 
        \label{fig:rf_regr_full_pdp_072718}
    \end{minipage}
\end{figure}

The performance of the RF regression model was first evaluated using data collected on individual flight dates. Given that weather conditions remain constant within a single flight date, the model inputs were reduced to thermal data and vegetation indices NDVI, NDRE, and PSRI. Figure \ref{fig:rf_regr_single_dates} presents the results of this analysis. Mean $R^{2}$ scores for the different flight dates were 0.08 (+/- 0.51) for 07/24/2017, -1.89 (+/- 3.40) for 08/22/2017, 0.17 (+/- 0.33) for 07/09/2018, and -0.03 (+/- 0.52) for 07/27/2018, respectively. These values indicate the model's varying levels of predictive accuracy on different dates. Furthermore, to illustrate the influence of each variable on SWP, partial dependence plots (PDPs) are presented in Figure \ref{fig:rf_regr_full_pdp_072417}, \ref{fig:rf_regr_full_pdp_082217}, \ref{fig:rf_regr_full_pdp_070918} and \ref{fig:rf_regr_full_pdp_072718}.

% The coefficient of determination ($R^{2})$) values for the different flight dates are as follows: 0.30 for 07/24/2017, 0.15 for 08/22/2017, 0.44 for 07/09/2018, and 0.17 for 07/27/2018. 


\subsection{Full RF Regression Model Based on Data from All Flight Dates}

The RF model was then assessed using aggregated data encompassing all flight dates. Due to considerable variations in weather parameters on different flight dates, it was essential to include these variations in the model to ensure comparability of thermal values and vegetation indices. Thus, the model's input variables comprised thermal data, vegetation indices (NDVI, NDRE, PSRI), and weather parameters (air temperature, VPD, and wind).

\begin{figure}
  % First row
  \begin{minipage}[c]{0.36\textwidth} % Adjust width as needed
    \centering
    \subfloat{\includegraphics[width=\textwidth]{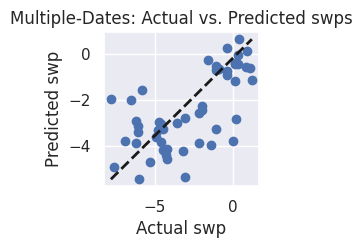}}
  \end{minipage}
  %\hfill % Space between the first and second subplot of the first row
  \hspace{4mm}
  \begin{minipage}[c]{0.34\textwidth} % Adjust width as needed
    \centering
    \subfloat{\includegraphics[width=\textwidth]{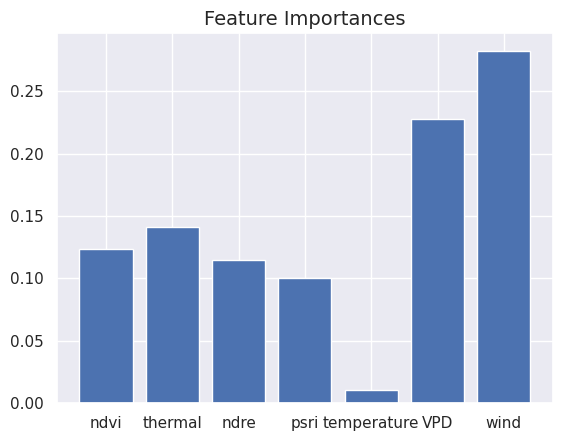}}\label{fig:rf_regr_var_importance}
  \end{minipage} 
  \caption{Performance of the Full Random Forest Regression Model Using Data from All Flight Dates, incorporating Weather Parameters. The accompanying variable importance plot ranks the factors in descending order of impact: wind, VPD, thermal data, NDVI, NDRE, PSRI, and air temperature.}
  \label{fig:rf_regr_alldates}
\end{figure}

\begin{figure}
    \centering
    \includegraphics[scale = 0.4]{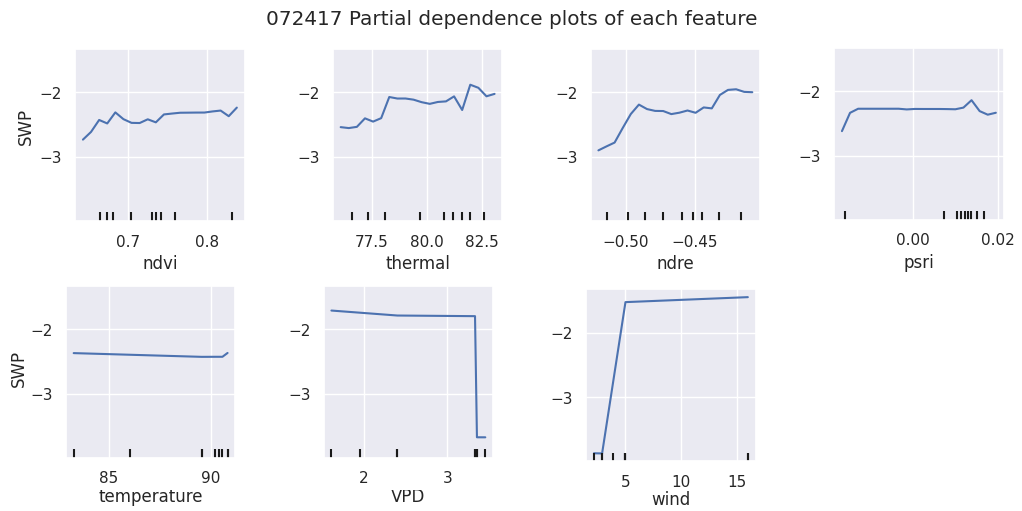}
    \caption{Partial dependence plots illustrating the influence of each feature on SWP measured across all of flight dates.} 
    \label{fig:rf_regr_full_pdp_alldates}
\end{figure}

As depicted in Figure \ref{fig:rf_regr_alldates}, this comprehensive approach resulted in a mean $R^{2}$ value of 0.65 (+/- 0.15), a significant increase compared to the single-date model results, highlighting a substantial enhancement in predictive accuracy. Furthermore, the right plot of the figure elucidates the variable importance ranking. It identifies the most impactful factors in descending order of influence: wind, VPD, thermal data, NDVI, NDRE, PSRI, and air temperature. This ranking offers valuable insights into the relative contribution of each variable to the performance of the model. Furthermore, to illustrate the influence of each variable on SWP, partial dependence plots (PDPs) are presented in Figure \ref{fig:rf_regr_full_pdp_alldates}.

\subsection{NoRedEdge RF Regression Model Based on Single Flight Date Data}

\begin{figure}
  \centering

  % Adjust each minipage width to fit four in a row
  \begin{minipage}[c]{0.24\textwidth}
    \centering
    \subfloat{\includegraphics[width=\textwidth]{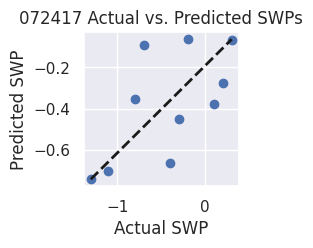}}\label{fig:rf_regr_norededge_072417}
  \end{minipage}
  \begin{minipage}[c]{0.24\textwidth}
    \centering
    \subfloat{\includegraphics[width=\textwidth]{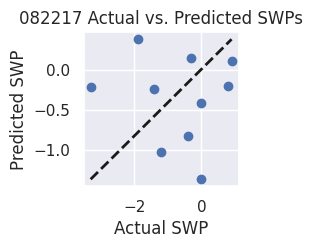}}\label{fig:rf_regr_norededge_082217}
  \end{minipage}
  \begin{minipage}[c]{0.24\textwidth}
    \centering
    \subfloat{\includegraphics[width=\textwidth]{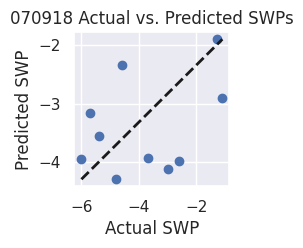}}\label{fig:rf_regr_norededge_070918}
  \end{minipage}
  \begin{minipage}[c]{0.24\textwidth}
    \centering
    \subfloat{\includegraphics[width=\textwidth]{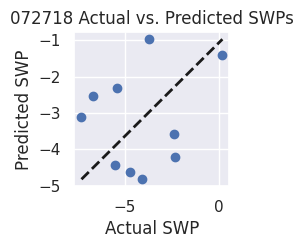}}\label{fig:rf_regr_norededge_072718}
  \end{minipage}

  \caption{Performance of the NoRedEdge Random Forest Regression Model: Predictions Using Data from Individual Flight Dates}
  \label{fig:rf_regr_norededge_single_dates}
\end{figure}

\begin{figure}[ht]
    \centering
    \begin{minipage}[t]{0.48\linewidth}
        \centering
        \includegraphics[scale = 0.56]{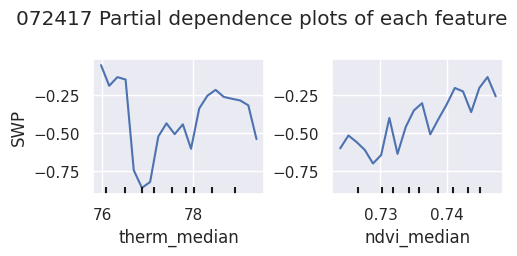}
        \caption{Partial dependence plots illustrating the influence of each feature on SWP measured on flight-date 072417.} 
        \label{fig:rf_regr_nore_pdp_072417}
    \end{minipage}\hfill
    \begin{minipage}[t]{0.48\linewidth}
       \centering
        \includegraphics[scale = 0.56]{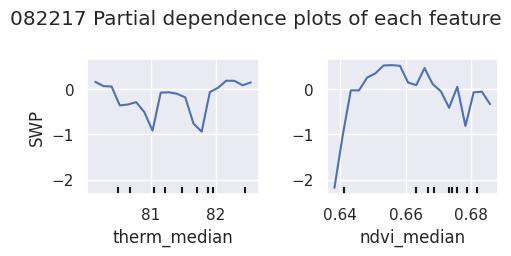}
        \caption{Partial dependence plots illustrating the influence of each feature on SWP measured on flight-date 082217.} 
        \label{fig:rf_regr_nore_pdp_082217}
    \end{minipage}
\end{figure}

\begin{figure}[ht]
    \centering
    \begin{minipage}[t]{0.48\linewidth}
        \centering
        \includegraphics[scale = 0.56]{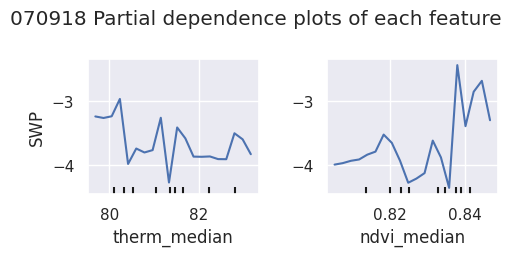}
        \caption{Partial dependence plots illustrating the influence of each feature on SWP measured on flight-date 070918.} 
        \label{fig:rf_regr_nore_pdp_070918}
    \end{minipage} \hfill
    \begin{minipage}[t]{0.48\linewidth}
        \centering
        \includegraphics[scale = 0.56]{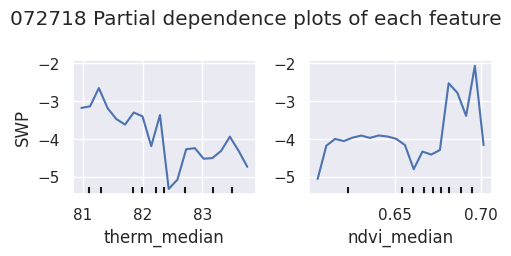}
        \caption{Partial dependence plots illustrating the influence of each feature on SWP measured on flight-date 072718.} 
        \label{fig:rf_regr_nore_pdp_072718}
    \end{minipage}
\end{figure}

The study progressed to assess a simplified RF model, specifically excluding the red-edge indices (NDRE and PSRI). This streamlined model, tailored for data from each individual flight date, incorporated only thermal data and NDVI as its input variables. The analysis results are presented in Figure \ref{fig:rf_regr_norededge_single_dates}. Mean $R^{2}$ scores for the different flight dates were -0.42 (+/- 0.87) for 07/24/2017, -3.50 (+/- 6.72) for 08/22/2017, -0.76 (+/- 1.42) for 07/09/2018, and -1.05 (+/- 1.06) for 07/27/2018, respectively. Comparatively lower than the $R^{2}$ values of the full RF model, these results demonstrate that excluding red-edge indices will reduce the model's predictive accuracy. Furthermore, to illustrate the influence of each variable on SWP, partial dependence plots (PDPs) are presented in Figure \ref{fig:rf_regr_nore_pdp_072417}, \ref{fig:rf_regr_nore_pdp_082217}, \ref{fig:rf_regr_nore_pdp_070918} and \ref{fig:rf_regr_nore_pdp_072718}.

% The coefficients of determination (R^{2}) for the different flight dates were 0.36 for 07/24/2017, -0.30 for 08/22/2017, 0.06 for 07/09/2018, and -0.36 for 07/27/2018, respectively. 


\subsection{NoRedEdge RF Regression Model Based on Data from All Flight Dates}

The study further explored the performance of the NoRedEdge RF regression model using aggregated data from all flight dates. Due to significant variations in weather factors across these dates, both the thermal values and vegetation indices exhibited considerable fluctuations. As such, a meaningful comparison of these variables necessitates the inclusion of weather factors in the model. Thus, the model inputs consisted of thermal data, NDVI, and weather parameters: VPD, wind, and air temperature.

\begin{figure}
  % First row
  \begin{minipage}[c]{0.36\textwidth} % Adjust width as needed
    \centering
    \subfloat{\includegraphics[width=\textwidth]{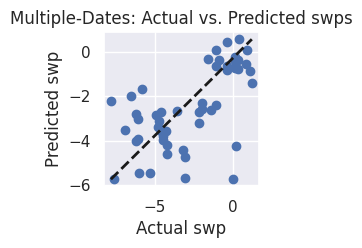}}
  \end{minipage}
  %\hfill % Space between the first and second subplot of the first row
   \hspace{4mm}
  \begin{minipage}[c]{0.34\textwidth} % Adjust width as needed
    \centering
    \subfloat{\includegraphics[width=\textwidth]{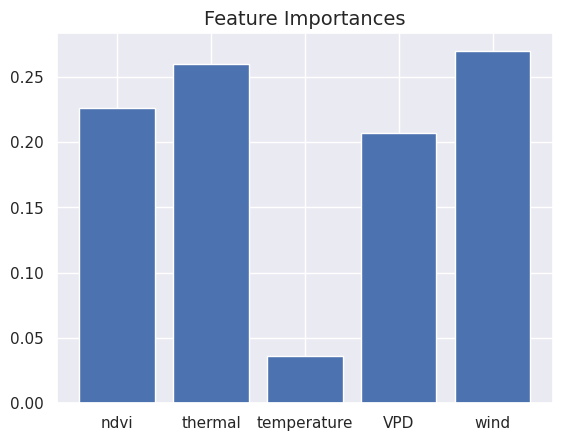}}\label{fig:rf_regr_norededge_var_importance}
  \end{minipage}
  \caption{Performance of the NoRedEdge RF Regression Model Using Data from All Flight Dates, incorporating Weather Parameters. The accompanying variable importance plot highlights the key factors in descending order of impact: wind, thermal data, NDVI, VPD, and air temperature.}
  \label{fig:rf_regr_norededge_alldates}
\end{figure}

\begin{figure}
    \centering
    \includegraphics[scale = 0.50]{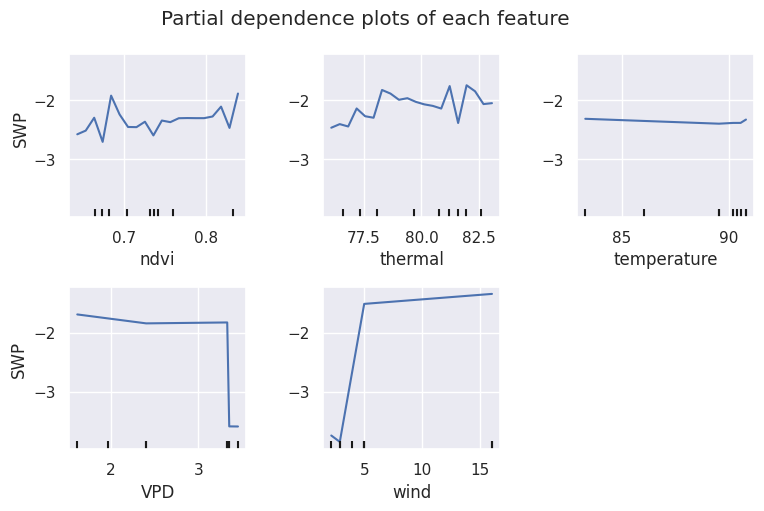}
    \caption{Partial dependence plots illustrating the influence of each feature on SWP measured across all of flight dates.} 
    \label{fig:rf_regr_norededge_pdp_alldates}
\end{figure}

The results, as shown in Figure \ref{fig:rf_regr_norededge_alldates}, reveal a mean $R^{2}$ score of 0.59 (+/- 0.17). This represents a notable improvement over the single-date models, yet it is marginally lower than the $R^{2}$ value of the full RF model. The right plot of the figure illustrates the variable importance, ranking the factors in descending order of influence: wind, thermal data, NDVI, VPD, and air temperature. This ranking provides insights into the contributions of these variables to the model's predictive capability.  Furthermore, to illustrate the influence of each variable on SWP, partial dependence plots (PDPs) are presented in Figure \ref{fig:rf_regr_norededge_pdp_alldates}.

\section{Water Pressure Classification Results}

\subsection{Full RF Classification Model Results}

\begin{figure}[ht]
  % First row
  \begin{minipage}[t]{0.33\textwidth} % Adjust width as needed
    \centering
    \subfloat{\includegraphics[width=\textwidth]{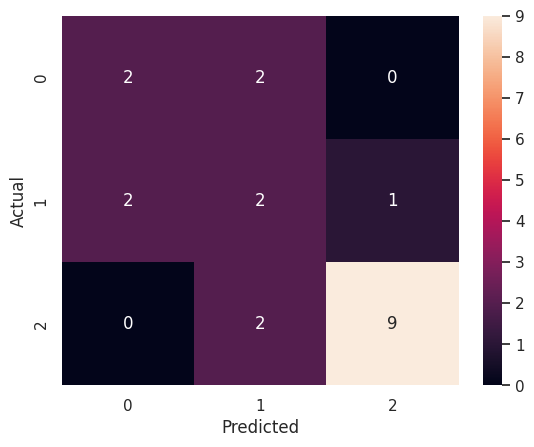}}\label{fig:rf_classify_confusion_matrix}
  \end{minipage}
  %\hfill % Space between the first and second subplot of the first row
  \begin{minipage}[t]{0.32\textwidth} % Adjust width as needed
    \centering
    \subfloat{\includegraphics[width=\textwidth]{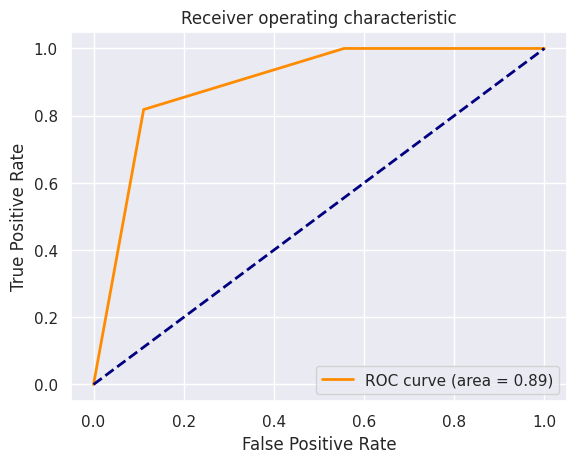}}\label{fig:rf_classify_roc}
  \end{minipage}
  \begin{minipage}[t]{0.33\textwidth} % Adjust width as needed
    \centering
    \subfloat{\includegraphics[width=\textwidth]{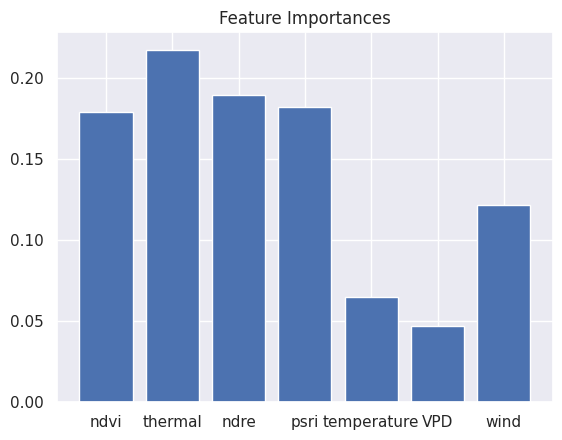}}\label{fig:rf_classify_var_importance}
  \end{minipage}
  \caption{Performance of the Full Random Forest Classification Model Using Data from All Flight Dates, Including Weather Parameters. The accompanying variable importance plot ranks the factors in descending order of impact: thermal data, NDRE, NDVI, PSRI, wind, air temperature, and VPD.}
  \label{fig:rf_classify_alldates}
\end{figure}

The comprehensive RF classification model, utilizing data from all flight dates, demonstrates notable performance metrics as illustrated in Figure \ref{fig:rf_classify_alldates}. The model achieved an accuracy of 85\% and an Area Under the Curve (AUC) of 0.86, indicating a high level of predictive capability. The figure's right plot details the variable importance, ranking the contributing factors in descending order of impact. These factors are, from most to least influential, thermal data, NDRE, NDVI, PSRI, wind, air temperature, and VPD.

\subsection{NoRedEdge RF Classification Model Results}

\begin{figure}[ht]
  \begin{minipage}[t]{0.33\textwidth} % Adjust width as needed
    \centering
    \subfloat{\includegraphics[width=\textwidth]{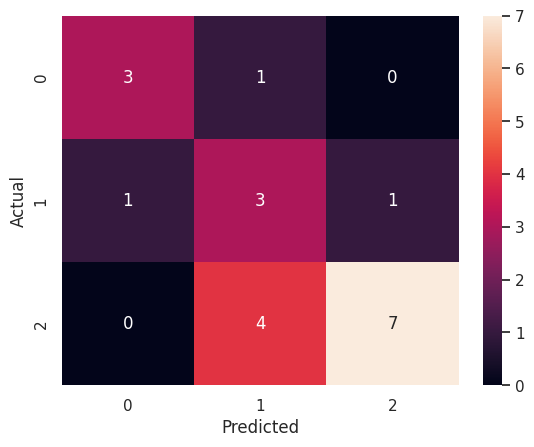}}\label{fig:rf_classify_norededge_confusion_matrix}
  \end{minipage}
  %\hfill % Space between the first and second subplot of the first row
  \begin{minipage}[t]{0.32\textwidth} % Adjust width as needed
    \centering
    \subfloat{\includegraphics[width=\textwidth]{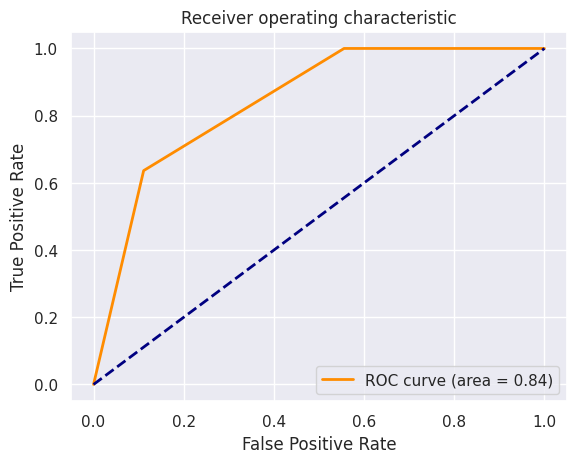}}\label{fig:rf_classify_norededge_roc}
  \end{minipage}
  \begin{minipage}[t]{0.33\textwidth} % Adjust width as needed
    \centering
    \subfloat{\includegraphics[width=\textwidth]{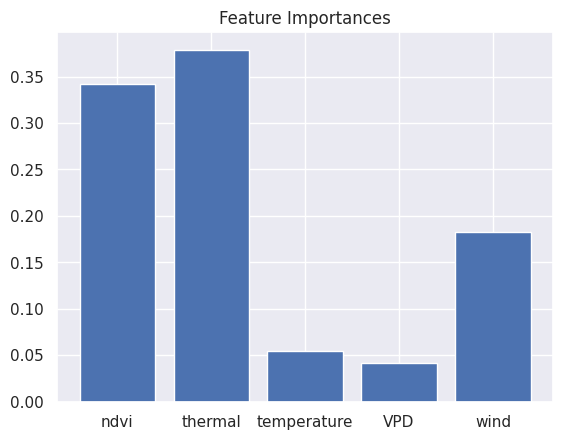}}\label{fig:rf_classify_norededge_var_importance}
  \end{minipage}
  \caption{Performance of the NoRedEdge RF Classification Model Utilizing Data from All Flight Dates, Including Weather Parameters. The accompanying variable importance plot ranks the key factors in descending order of impact: thermal data, NDVI, wind, air temperature, and VPD.}
  \label{fig:rf_classify_norededge_alldates}
\end{figure}

The NoRedEdge RF classification model, specifically designed to exclude red-edge indices while incorporating data from all flight dates, is depicted in Figure \ref{fig:rf_classify_norededge_alldates}. This model achieved an accuracy of 80\% and an Area Under the Curve (AUC) of 0.86. These results signify a competent level of predictive capability, albeit slightly diminished due to the absence of red-edge input variables. The figure's right plot delineates the variable importance, methodically ranking the most influential factors in descending order: thermal data, NDVI, wind, air temperature, and VPD.

\section{Discussion}

\subsection{Relationship Between SWP, Vegetation Indices, and Weather Variables}

In the initial phase of our study, we employed statistical linear regression to examine the univariable correlations between field-measured SWP and individual variables, such as thermal data and vegetation indices (VIs). The results from this approach did not reveal strong correlations, suggesting a more complex relationship between SWP and these variables. Additionally, we observed significant variations in SWP, thermal data, and VIs across different UAV flight dates, which indicates the need to incorporate environmental and weather variations when comparing and aggregating these measurements.

We further expanded our study by employing a machine-learning approach to analyze the multivariable relationship between field-measured SWPs and various environmental and spectral factors. This methodology, which incorporates a broader range of variables, achieved good results. 

In our research, the RF model utilized raw thermal band readings without converting them to Crop Water Stress Index (CWSI). This approach was justified as CWSI is a linear transformation of thermal value and RF model was observed to manage it well. Future studies could explore whether using CWSI as an input instead of raw thermal data further enhances model performance.

The weather factors, wind, air temperature, and VPD are essential to aggregate data from different flight dates, especially wind, which is shown as the most significant weather factor in the SWP regression models.

NDVI is shown as the most significant vegetation index in our models, followed by the thermal band, NDRE, and PSRI. The inclusion of red-edge indices NDRE and PSRI, further enhanced SWP mapping accuracy. Future research will explore additional vegetation indices to be used in the RF models to identify the most impactful set, aiming to further improve SWP estimation performance.

\subsection{Error and Uncertainty Analysis}
Various factors contribute to errors in accurately capturing canopy features. These include disruption from background soil, canopy shadows, neighboring canopies, per-canopy environmental conditions such as wind, and the orchard's topography.

Canopy shadows can distort thermal values in UAV imagery. To mitigate this, measurements were primarily conducted around noon to minimize shadow effects. However, fluctuations in data due to shadows were still observed. To tackle this issue, we used the Normalized Excess Green Index (NExG) as a filtering tool to isolate and remove dark shadows from the green canopy in our analysis.

In our study, weather data were obtained from a weather station nearby, which may not have precisely reflected the specific local conditions of the orchard. Furthermore, we applied these weather variables uniformly across all sampled walnut trees. This approach lacked precision, as it did not consider variations in wind speed and air temperature throughout different parts of the orchard. Notably, we observed that the irrigation blocks located in the upper left corner consistently showed higher thermal readings than those at the lower bottom edge. For future study, implementing more localized weather tracking within the orchard would be beneficial to further enhance the accuracy of water stress estimation.

\section{Conclusion and Future Directions}

\subsection{Conclusion}
This study delves into mapping stem water potential (SWP) in walnut orchards using multispectral UAV remote sensing data, aiming to provide a scalable and cost-effective tool for precision irrigation management at the individual plant level. We explored various vegetation indices relevant to plant water stress, including infrared NDVI, red-edge NDRE, and PSRI, along with thermal-band readings and weather data such as wind, air temperature, and VPD.

Utilizing the random forest algorithm, we developed two models: one for SWP regression and another for SWP classification. Both models performed effectively, achieving an R2 value of 0.8 and an accuracy of 85\%, respectively. Significant variables in SWP estimation included wind, NDVI, thermal, NDRE, and PSRI, in descending order. These results validate the effectiveness of UAV-based multispectral imaging and machine learning in assessing walnut water stress, integrating thermal data, NDVI, red-edge indices, and weather variables.

The approach is promising as it can replace manual, labor-intensive measurements with precise, targeted irrigation management at the individual plant level, marking a substantial step forward in refining and optimizing irrigation strategies.

\subsection{Future Work}
For future research, an extension of our study involves estimating SWPs for each walnut tree across the entire orchard. These individual tree-level results would then be aggregated based on irrigation blocks. This approach would allow us to examine patterns at the block level, providing insights into spatial and temporal variations. It will also reveal trends or anomalies in water stress across different areas of the orchard over time, contributing to a more nuanced understanding of orchard-wide water management needs.

\section*{Acknowledgements}
This project owes its success to the invaluable guidance and supervision of Dr. Yufang Jin, Minmeng Tang, and Zhehan Tang, whose contributions were indispensable. Additionally, heartfelt thanks are extended to all of the researchers in the Jin Lab at UC Davis, with whom I had the privilege of conducting field observations. The collaboration and support from the team were instrumental in the completion of this project.

\section*{Data Availability}
Availability of data The data that support the findings of this study are available from the corresponding author, upon reasonable request. 

\section*{Code Availability}
The code that was used for data analysis in the current study is available on request from the corresponding author.

\section*{Declarations: Conflict of interest}
The authors declare no conflict of interest.

%\clearpage
\bibliography{refs}
%\bibliographystyle{mnras}
%\bibliographystyle{mnras}

%\end{acknowledgments}

%\clearpage

% Don't change these lines
%\bsp	% typesetting comment
\label{lastpage}

\end{document}